\documentclass[11pt,a4paper]{article}
\usepackage[hyperref]{acl2020}
\usepackage{times}
\usepackage{latexsym}
\usepackage{verbatim}
\usepackage{xspace}
\usepackage{booktabs}
\usepackage{graphicx}
\usepackage{amsthm}
\usepackage{amsfonts}
\usepackage{mathtools}
\usepackage{url}

\aclfinalcopy

\title{
Don't Say {\em That}!\\ Making Inconsistent Dialogue Unlikely with Unlikelihood Training
}
\author{
  Margaret Li$^{1}$, Stephen Roller$^{1}$, Ilia Kulikov$^{2\star}$, Sean Welleck$^{2\star}$\\ 
  \textbf{Y-Lan Boureau$^{1}$, Kyunghyun Cho$^{1,2}$, Jason Weston$^{1,2}$}\\
  $^1$Facebook AI Research \\
  $^2$New York University \\
  \texttt{\{margaretli,roller,ylan,kyunghyuncho,jase\}@fb.com} \\
  \texttt{wellecks@nyu.edu,kulikov@cs.nyu.edu}
}

\date{}

\begin{document}
\maketitle

\begin{abstract}
Generative dialogue models currently suffer from a number of problems which standard maximum likelihood training does not address. They tend to produce generations that (i) rely too much on copying from the context, (ii) contain repetitions within utterances, (iii) overuse frequent words, and (iv) at a deeper level, contain logical flaws. In this work we show how all of these problems can be addressed by extending the recently introduced unlikelihood loss \citep{welleck2019neural} to these cases. We show that appropriate loss functions which regularize generated outputs to match human distributions are effective for the first three issues.  For the last important general issue, we show applying unlikelihood to  collected data of {\em what a model should not do}  is effective for improving logical consistency, potentially paving the way to generative models with greater reasoning ability. We demonstrate the efficacy of our approach across several dialogue tasks.
\end{abstract}

\let\thefootnote\relax\footnotetext{$^{\star}$Work done while at Facebook AI Research (FAIR).}

\section{Introduction}
\begin{figure}[t!]
    \centering
    \includegraphics[width=\linewidth]{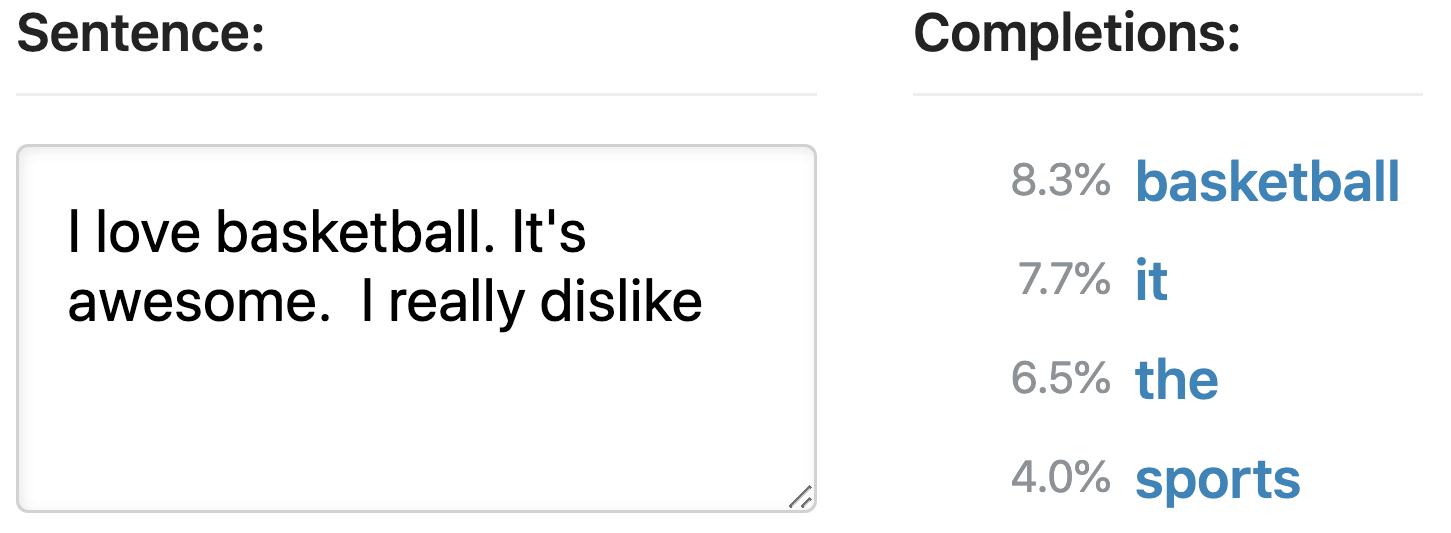}
    \caption{GPT-2 345M model completions can show lack of coherence, e.g. direct contradictions.}
    \label{fig:gpt2_contra}
\end{figure}

Open-ended tasks such as dialogue reveal a number of issues with current neural text generation methods.
In more strongly grounded tasks such as machine translation
and image captioning, current encoder-decoder architectures
provide strong performance, where mostly word-level decisions
are often taken correctly by the model.
However, 
critical failings are exposed  in less constrained generation: 
 reliance on repetitive copying and overuse of frequent words,
and an inability to maintain logical coherence.
The former shows the learning objective is faulty in that
it cannot match simple statistics of the training data,
while the latter touches more to the heart of artificial
intelligence: these models do not understand what they are saying.
For example, Figure \ref{fig:gpt2_contra} shows how the 345M-parameter GPT2 model \citep{radford2019language} can give high probability to contradictory generations.

In this work, we show how the recently introduced unlikelihood
objective \citep{welleck2019neural} can be generalized
to remedy these problems.
Unlikelihood is a technique developed for removal of repetition in language model
completions, and works by adding an extra term to the objective that forces
repetitions to have low probability, alleviating the degenerative problems 
highlighted in \citet{holtzman2019curious}. In fact, unlikelihood can be seen as a much more general framework, as we will see.

We first generalize unlikelihood to a different domain: dialogue, where 
we measure statistics of the training distribution in terms of contextual copies, 
within-utterance repeats, and vocabulary usage. We then develop loss functions
that control these statistics, providing improved metrics on several tasks.
Secondly, we show how the same tools can be used to address deeper semantic issues
in such models. By leveraging existing natural language inference (NLI)
data  \citep{welleck2018dialogue} as supervision against poor quality 
generations,
we train models that assign low probability to generating incoherent and
contradictory
text. Overall, our  approach  yields more consistent dialogue models across several axes, and provides a promising framework for further advances.

Code and pre-trained models will be made available.$^{\dagger}$\footnote{$^{\dagger}$\url{https://parl.ai/projects/dialogue_unlikelihood/}}

\section{Dialogue Unlikelihood Training}
\label{sec:ul-training}
\paragraph{Dialogue Generation} Dialogue generation consists in predicting an utterance $\mathbf{y}=(y_1,\ldots,y_{|y|})$ given a context $\mathbf{x}=\{s_1,\ldots,s_k,u_1,\ldots,u_t\}$ that consists of initial context sentences $s_{1:k}$ (e.g., scenario, knowledge, personas, etc.) followed by dialogue history utterances  $u_{1:t}$  from speakers who  take  consecutive turns.
\paragraph{Likelihood Training} Given a dataset $\mathcal{D}=\{(\mathbf{x}^{(i)},\mathbf{y}^{(i)})\}$ derived from a collection of human-human interactions, the standard approach to generative training for dialogue tasks is
maximum likelihood estimation (MLE), that minimizes:
\fontsize{10}{12}{
\begin{equation*}
\mathcal{L}^{(i)}_{\text{MLE}}(p_{\theta}, \mathbf{x}^{(i)}, \mathbf{y}^{(i)})=-\sum_{t=1}^{|y^{(i)}|}\log p_{\theta}(y^{(i)}_t|\mathbf{x}^{(i)}, y^{(i)}_{<t}),
\end{equation*}
}where  $\mathbf{x}^{(i)}$ is a gold context (dialogue history and initial context sentences) and $\mathbf{y}^{(i)}$ is a gold next-utterance, and $y^{(i)}_t$ is the $t$-th token of $\mathbf{y}^{(i)}$.

Likelihood-based (greedy or beam) decoding applied after training a model with this objective yields sequences with statistics that do not match the original human training sequence distribution.

\paragraph{Unlikelihood Training} To control for such distribution mismatches, we employ  the unlikelihood loss 
\citep{welleck2019neural}, generalizing it to our setting, and developing a particular form of the loss function for each type of mismatch. 

The general form of the unlikelihood loss penalizes a set of tokens $\mathcal{C}_{t}$ at each time-step, $\mathcal{L}^{(i)}_{\text{UL}}(p_{\theta}, \mathcal{C}_{1:T}, \mathbf{x}, \mathbf{y})=$
\begin{align*}
     -\sum_{t=1}^{|y|}\sum_{y_{c}\in \mathcal{C}_t}\beta(y_c)\log \left(1-p_{\theta}(y_{c}|\mathbf{x},y_{<t})\right),
\end{align*}
where $\mathcal{C}_t\subseteq \mathcal{V}$ is a subset of the vocabulary, and $\beta(y_c)$ is a candidate-dependent scale that controls how much the candidate token should be penalized.
The overall objective in
unlikelihood training then consists of mixing the likelihood and unlikelihood losses,
\begin{equation}\label{eq:mix}
    \mathcal{L}^{(i)}_{\text{ULE}}=\mathcal{L}^{(i)}_{\text{MLE}}+\alpha\mathcal{L}^{(i)}_{\text{UL}},
\end{equation}
where $\alpha\in\mathbb{R}$ is the mixing hyper-parameter.

Likelihood tries to model the overall sequence probability distribution, while
unlikelihood corrects for known biases.
It does this via the set of {\em negative candidates} $\mathcal{C}_t$ calculated at each step $t$, where we are free to select candidate generation functions
depending on the biases to be mitigated.
Likelihood pushes {\em up} the probability of a gold token $y^{(i)}_t$
while unlikelihood pushes {\em down} the probability of negative candidate tokens $y_c \in \mathcal{C}_t$.

In \citet{welleck2019neural} the context $\mathbf{x}$ consists of a ground-truth sequence ($\mathbf{x}=\mathbf{x}^{(i)}$),
the target $\mathbf{y}$ is either a ground-truth sequence ($\mathbf{y}=\mathbf{y}^{(i)}$) or a model-generated sequence ($\mathbf{y}=\hat{\mathbf{y}}$),  and the per-token scale parameter $\beta(y_c)$ is $1$.

In this paper, we demonstrate how unlikelihood can be used as a general framework by applying it to the dialogue domain. We show how varying the contexts $\mathbf{x}$, targets $\mathbf{y}$, candidates $\mathcal{C}$ and scaling $\beta$ can be used to improve the coherence and language modeling quality of dialogue models. 
To do this,
we now consider the different biases we wish to mitigate, and construct a specific
unlikelihood  loss for each in turn.

\subsection{Repetition and Copying}
\label{ssec:rep-ul}
Generative dialogue models are known to both (i) rely too much on copying existing context knowledge or dialogue history; and (ii) repeat themselves within individual utterances.
To address this with unlikelihood, we define two types of negative candidate
tokens which either appear in a repeating n-gram from the context or from the generated label itself,
\begin{align*}
    \mathcal{C}_t^{\text{context-copy}}=& \begin{cases}\{y_t\} & y_t\in \text{{\small repeat context n-gram}}\\ \emptyset & \text{\small otherwise},\end{cases}\\
    \mathcal{C}_t^{\text{label-repeat}}=& \begin{cases}\{y_t\} & y_t\in \text{\small repeating label n-gram}\\ \emptyset & \text{\small otherwise},\end{cases}
\end{align*}
where $y_t$ is a token in a repeating context n-gram when $y_t$ is part of an n-gram that already appeared in the context tokens $x$, and is in a repeating label n-gram when $y_t$ is part of an n-gram that already appeared in $y_{<t}$.
 Given a ground-truth context $\mathbf{x}^{(i)}$, we apply these two forms of unlikelihood to a model-generated sequence $\hat{\mathbf{y}}^{(i)}$.
 In summary, we either apply the per-example loss 
 \begin{align*}
 \mathcal{L}_{\text{UL}}^{(i)}(p_{\theta}, \mathcal{C}_{1:|y|}^{\text{context-copy}},\mathbf{x}^{(i)},\hat{\mathbf{y}}^{(i)})
 \end{align*}
 for controlling context copies, or  
 \begin{align*}
  \mathcal{L}_{\text{UL}}^{(i)}(p_{\theta}, \mathcal{C}_{1:|y|}^{\text{label-repeat}},\mathbf{x}^{(i)},\hat{\mathbf{y}}^{(i)}).
  \end{align*}
  for controlling label repeats.
  We also consider mixing the two losses to mitigate both issues.

\subsection{Vocabulary Usage}
\label{ssec:vocab-ul}
Neural sequence models trained with maximum likelihood generate sequences with token distributions that differ from those of human text \cite{dinan2019second,holtzman2019curious}. 
In particular, these models tend to produce high frequency tokens too often and low frequency tokens too rarely, where frequency is defined by the human token distribution. 

We address this with unlikelihood by penalizing tokens according to the mismatch between the model and ground-truth unigram distributions. Specifically, we first maintain an empirical estimate of the model's unigram distribution $p_{\text{model}}(y_t)$ and the human distribution $p_*(y_t)$:
\begin{align*}
    p_{\text{model}}(y_t)&=\frac{\text{count}(y_t)}{|Y|},
\end{align*}
where $Y$ is a collection of token predictions on a subset of training data $\mathcal{D}'$ (e.g. the preceding $k~=~256$ batches), and $\text{count}(y_t)$ is the number of occurrences of 
$y_t$ in $Y$. 
This is  computed using model sequences $(\mathbf{y}=\hat{\mathbf{y}})$, defining $Y$ as the collection of all tokens in all $\hat{\mathbf{y}}$.

We wish to {\em push down} the probability of tokens appearing too often, i.e. when 
 $p_{\text{model}}(y_t) > p_*(y_t)$.
For the unlikelihood loss, 
each step's candidate is thus the current token, $\mathcal{C}_t^{\text{identity}}=\{y_t\}$,
and each token's unlikelihood loss is scaled according to the mismatch between the approximated model and human distributions, 
\begin{align*}
    \beta(y_c) = p_{\text{model}(y_c)}\log\left(\frac{p_{\text{model}}(y_c)}{p_{*}(y_c)}\right).
\end{align*}
The unlikelihood loss for a token $y_c$ is non-zero when the token occurs more often in the model's estimated unigram distribution. In summary, the resulting per-example loss is
\begin{align*}
    \mathcal{L}^{(i)}_{\text{UL}}(p_{\theta}, \mathcal{C}_{1:|y|}^{\text{identity}},\mathbf{x}^{(i)},\mathbf{y})
\end{align*}where $\mathbf{y}$ is a model-generated sequence.

\subsection{Contradictions}
\label{ssec:contra-ul}
Neural generation models appear fluent, especially when pre-trained on large datasets,
but are still poor at understanding the language they produce. That is, they can produce 
logically or factually inaccurate, or contradicting statements \citep{welleck2018dialogue,zhang2018personalizing,hayashi2019latent,petroni2019language}.
Here,  we show how the unlikelihood objective can be used to train such models to assign low probability to inconsistent and contradictory utterances. 

To do so, we assume the existence of training data of both positive {\em and} negative examples of coherent behavior. There is a raft of recent large-scale, high quality data that can be massaged into this form,  from natural language inference (NLI) tasks \citep{Bowman2015snli,Williams2017mnli,welleck2018dialogue} to
commonsense reasoning tasks \citep{zellers2019hellaswag,qin2019counterfactual}. 
Two collections of data can be derived from the labels 
of such a supervised task:
\begin{equation*}
{\cal D}^{+} = \{ (\mathbf{x}^{(i)},\mathbf{y}^{(i) +}) \},~~~
{\cal D}^{-} = \{ (\mathbf{x}^{(i)},\mathbf{y}^{(i) -}) \},
\end{equation*}
 where ${\cal D}^{+}$
is coherent behavior, e.g. neutral or entailing data in NLI, and  ${\cal D}^{-}$
is incoherent behavior, e.g. contradictions. In general, many forms of this 
type of data can be collected, not just NLI, and it is also not necessary for the contexts $\mathbf{x}^{(i)}$ to overlap as we have written here.

Standard likelihood training can then be performed on coherent data   ${\cal D}^{+}$,
while the unlikelihood objective
is applied to ${\cal D}^{-}$ as we wish to {\em push down} the probability of generating
the  incoherent response $\mathbf{y}^{-}$ given a context $\mathbf{x}$.
That is, given an incoherent pair  $(\mathbf{x}, \mathbf{y}^{-})$
we use the loss
\begin{align*}
    \mathcal{L}_{\text{UL}}(p_{\theta},\mathcal{C}_{1:|y|}^{\text{identity}},\mathbf{x}, \mathbf{y}^{-}),
\end{align*}
where we penalize each token in the target ($\mathcal{C}_t^{\text{identity}}=\{y^{-}_t\}$). Hence, the loss makes generating the contradicting sentences less likely.

\section{Related Work}

Our work provides new applications of unlikelihood training \citep{welleck2019neural}, showing that  unlikelihood offers
a general framework for improving generative models, and in particular dialogue models.
Outside of that work,
the use of negative training in	dialogue retrieval, rather than generation, has been previously extensively studied,
see e.g. \citep{humeau2019real,nugmanova2019strategy}.
In the area of generative dialogue, a number of works have focused on improving the standard likelihood training approach.
Closer to our work is that of \citet{he2019negative}  which 
developed the approach of negative training
to prevent generic and malicious responses in dialogue models. 
In terms of improving repetition and specificity, a recent alternative approach 
is that of control 
\citep{fan2018controllable, ficler2017controlling,ghazvininejad2017hafez,see2019makes}.
Nucleus sampling \cite{holtzman2019curious} can help to remove generic or repetitive utterances at the expense of accuracy, but was shown to be inferior to beam blocking, which in turn was shown to be inferior to unlikelihood in \citet{welleck2019neural}.

In terms of dialogue coherence,
 \citet{welleck2018dialogue} showed that retrieval, but not generative models, could be improved with NLI as a re-scorer, while \citet{yang2018learning} 
multi-tasked with NLI. 
The work of \citet{gabriel2019cooperative} has also studied improving narrative flow with a discriminative rescorer,  but in that case for generated language.
In our work, the improvements are
tightly integrated into the training of the model itself.

\section{Experiments}

In all of our experiments we employ a large pre-trained seq2seq Transformer \cite{vaswani2017attention} as our base 
model, which we then fine-tune for particular tasks with the objectives outlined in Section~\ref{sec:ul-training} and specified in each experiment below.
Following previous work \citep{humeau2019real}, we pre-train our model on
dialogue data, using a previously existing Reddit dataset extracted and obtained by a third party and made available on pushshift.io, training to generate a comment conditioned on the full thread leading up to the comment, spanning $\sim 2200M$ training examples. 
Our Transformer model consists of an 8 layer encoder, 8 layer decoder with 512-dimensional embeddings and 16 attention heads, and is based on the ParlAI implementation of \citet{miller2017parlai}.
The model was trained with a batch size of 3072 sequences for approximately 3M updates using a learning rate of 5e-4, and an inverse square root scheduler. This pre-training took approximately two weeks using 64 NVIDIA V100s.

\begin{table}[t!]
\setlength{\tabcolsep}{4pt}
    \centering
    \resizebox{\linewidth}{!}{
    \begin{tabular}{lrrrr}
\toprule
     &     &     & \multicolumn{2}{c}{Repetition} \\
     \cmidrule(lr){4-5}
 Model    & PPL   & F1   & Context & Label \\
\midrule
Human   &  - & -  & .0223   & .0004 \\
MLE Baseline             & 11.4 & .199   & .1131 & .0210 \\
 \addlinespace[0.3em]
UL (Context only)      &    11.8 &    .194   &     .0330 &     .0069 \\
UL (Label only)        &    11.4 &    .203   &     .0984 &     .0005 \\
UL (Context \& Label)  &    11.9  &   .193   &     .0352 &     .0023 \\
\bottomrule
\end{tabular}
    }
    \caption{Evaluation on the ConvAI2 task valid set (test set is hidden), comparing standard likelihood (MLE) with context and label repetition unlikelihood loss training. 
The repetition types can be decreased depending on which type of unlikelihood loss is used,  with minimal changes in perplexity and F1.
 }
    \label{tab:rep_convai}
    \vspace{5mm}
\setlength{\tabcolsep}{4pt}
    \centering
    \resizebox{0.95\linewidth}{!}{
    \begin{tabular}{lrrrr}
\toprule
     &     &   &  \multicolumn{2}{c}{Repetition} \\
     \cmidrule(lr){4-5}
 Model         & PPL  & F1  & Context & Label \\
\midrule
Human                 & -    &  -   & .160 & .001  \\
MLE Baseline          & 8.3  & .368 & .441 & .014 \\
\addlinespace[0.3em]
UL (Context only)      & 8.8  & .346 &  .229  & .037 \\
UL (Label only)        & 8.3  & .371 &  .426  & .001 \\
UL (Context + Label)  & 8.5  & .358 &  .313  & .009 \\
\bottomrule    
\end{tabular}
    }
    \caption{Evaluation on the Wizard of Wikipedia test set, comparing standard likelihood (MLE)
 with context and label repetition unlikelihood loss training. The repetition types can be decreased depending on the type of unlikelihood loss used, while minimally impacting F1.}
    \label{tab:rep_wiz}
\end{table}

\begin{table}[h]
\setlength{\tabcolsep}{4pt}
    \centering
    \resizebox{0.95\linewidth}{!}{
    \begin{tabular}{lrrrrr}
\toprule
     &     &   &   \multicolumn{2}{c}{Repetition} \\
     \cmidrule(lr){4-5}
  Model         & PPL  & F1   &  Context & Label \\
\midrule
Human                 & -    &  -   &    .009 & .010 \\
MLE Baseline          & 21.0 & .130 &    .033 & .617 \\
\addlinespace[0.3em]
UL (Context only)    & 21.4 & .163 &  .008  & .322 \\
UL (Label only)      & 21.4 & .183 &  .015  & .055 \\
UL (Context + Label) & 21.8 & .184 &  .009  & .078 \\
\addlinespace[0.3em] 
\bottomrule    
\end{tabular}
    }
    \caption{Evaluation on the ELI5 task test set, comparing standard likelihood (MLE)
 with context and label repetition unlikelihood loss training.
The repetition types can be decreased depending on which type of unlikelihood loss is used, while improving F1.}
    \label{tab:rep_eli5}
\end{table}

\subsection{Repetition and Copying}
We use the ConvAI2 persona-based dialogue \citep{zhang2018personalizing},
Wizard of Wikipedia knowledge-grounded dialogue \cite{dinan2018wizard}
and ELI5 long-form question answering \citep{fan2019eli5} datasets 
to evaluate the effect of using unlikelihood to reduce copying and repetition in model generated utterances. 
On each dataset, we fine-tune the pre-trained pushshift.io Reddit model, then evaluate by generating next-utterances for dialogue contexts from the test set (or validation in ConvAI2, as the test set is hidden). 
We use greedy decoding in our main experiments for simplicity and scalability,
but we also obtained similar results with beam search, shown in Appendix
\ref{app:beam}.

To measure label repetition in a sequence $\textbf{y}$, we use the portion of duplicate n-grams:
\begin{align*}
    1.0-\frac{|\text{unique n-grams}(\textbf{y})|}{|\text{n-grams}(\textbf{y})|},
\end{align*}
and report the metric averaged over the examples. Label repetition increases from zero as the model generates more repeated n-grams.
To measure context repetition, we measure the fraction of generated n-grams  that appear in the original context:
\begin{align*}
    \frac{|\text{n-grams}(\textbf{y}) \cap \text{n-grams}(\textbf{x})|}{|\text{n-grams}(\textbf{y})|},
\end{align*}
and report the metric averaged over the examples. Context repetition increases when the model `copies' n-grams from the context. To quantify 
language modeling quality, we use standard perplexity and \textsc{F1} metrics. 

We use the pre-trained model fine-tuned with MLE as the baseline, and compare it against the pre-trained model fine-tuned with copy and repetition unlikelihood (\S\ref{ssec:rep-ul}).

\paragraph{Results}

Results for ConvAI2 are shown in Table~\ref{tab:rep_convai}. We see that training unlikelihood using only-contexts or only-labels reduces their corresponding metrics dramatically compared to the MLE baseline. Training with both context- and label-repetition unlikelihood reduced both context repetitions (by 69\%, .0352 vs. .1131) and label repetitions (by 89\%, .0023 vs .0210) compared to the MLE baseline, much closer to human levels, while keeping perplexity essentially constant.

Comparatively, the Wizard of Wikipedia MLE baseline experiences a much larger problem with context repetition, due to its tendency to copy grounded knowledge verbatim (Table~\ref{tab:rep_wiz}). 

Results for ELI5, shown in Table~\ref{tab:rep_eli5}, show that
it has an especially large problem with label repetition, and that label-unlikelihood is able to reduce the repetitions by 91\% (.055 vs .617), while significantly boosting F1 (.130 to .182).

\begin{figure}[t]
    \centering
    \includegraphics[width=\linewidth]{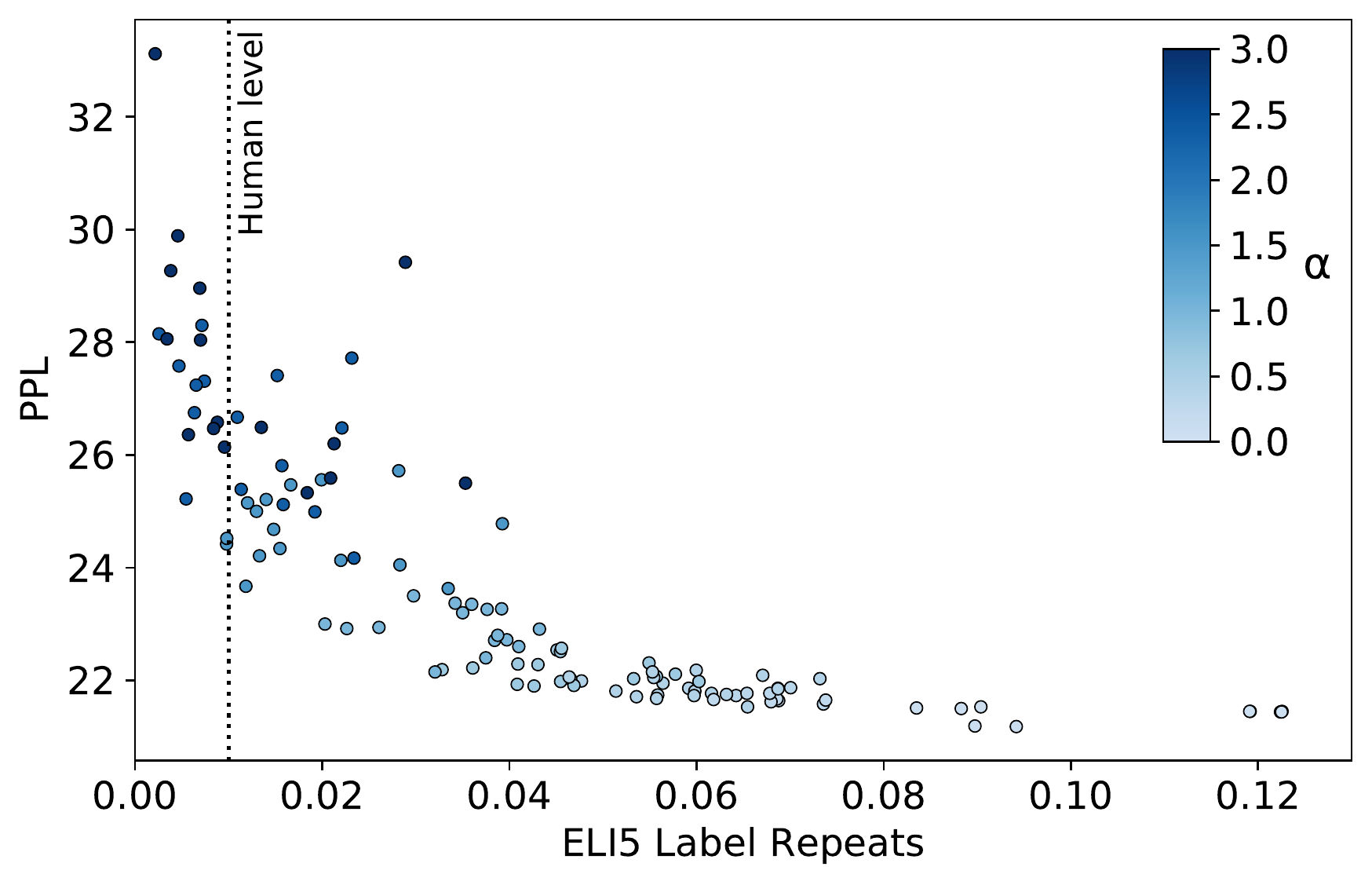}
    \caption{ELI5: Perplexity vs. label repeats as a function of $\alpha$ in the label unlikelihood objective.  }
    \label{fig:eli5label}
\vspace{5mm}
    \centering
    \includegraphics[width=\linewidth]{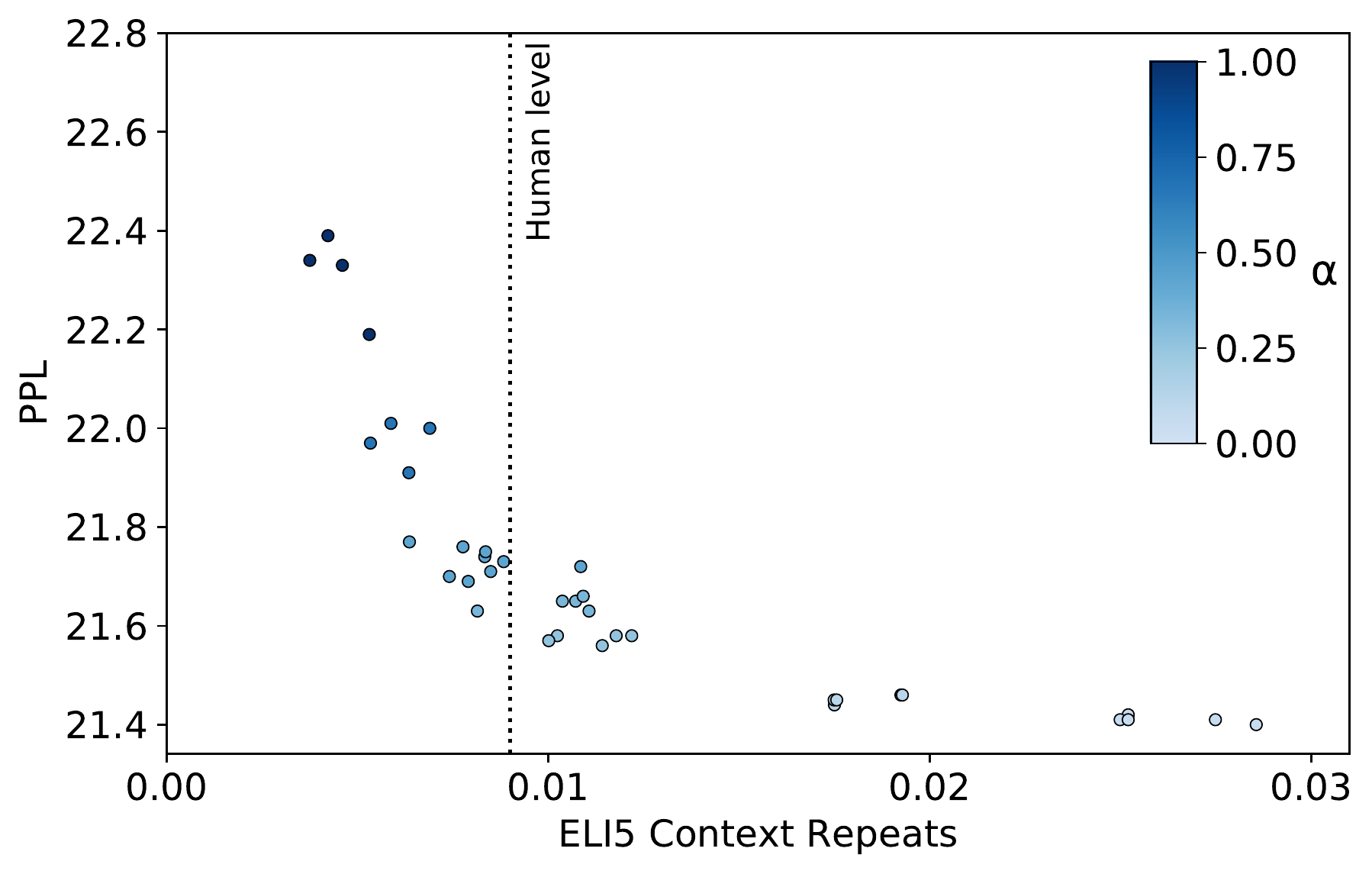}
    \caption{ELI5: Perplexity vs. context repeats as a function of $\alpha$ in the context unlikelihood objective. }
    \label{fig:eli5context}
\end{figure}

Figures \ref{fig:eli5label} and \ref{fig:eli5context}
show perplexity as a function of label and context repeats respectively using unlikelihood on ELI5. The parameter $\alpha$ can clearly control repeats smoothly, with only very high values resulting in increased perplexity.

\paragraph{Human Evaluation}

Finally, we perform a human evaluation using the same pairwise evaluation scheme as \citep{fan2019eli5} performed on ELI5, comparing the MLE baseline 
to UL (Label only) which asks: {\em Which response answers the question better?} The evaluators are asked to consider both the readability and accuracy of the answer. Results are given in Figure~\ref{fig:human_eval} (left), 
showing a statistically significant improvement over the baseline 
(150 trials, two tailed binomial test, $p<0.01$). 
Further details are given in Appendix \ref{app:human_eval}.

\subsection{Vocabulary Usage}
We evaluate the ability of vocabulary unlikelihood (\S\ref{ssec:vocab-ul}) to reduce the mismatch between model and human token distributions. 

We use the ConvAI2 dataset, where our 
baseline is again trained using maximum likelihood.
Starting with the baseline model, we then fine-tune several models using vocab unlikelihood at logarithmically interpolated values of $\alpha \in [1, 1000]$.

We partition the vocabulary into `frequent', `medium', `rare', and `rarest' using the human unigram distribution computed with the ConvAI2 training set, corresponding to the sorted token sets whose cumulative mass accounts for the top 40\%, the next 30\%, the next 20\% and the final 10\% of usage, respectively. We evaluate a model by generating utterances given contexts from the ConvAI2 validation set, and compute the fraction of tokens within each class.

\paragraph{Results}

\begin{figure}[t!]
    \centering
    \includegraphics[width=\linewidth]{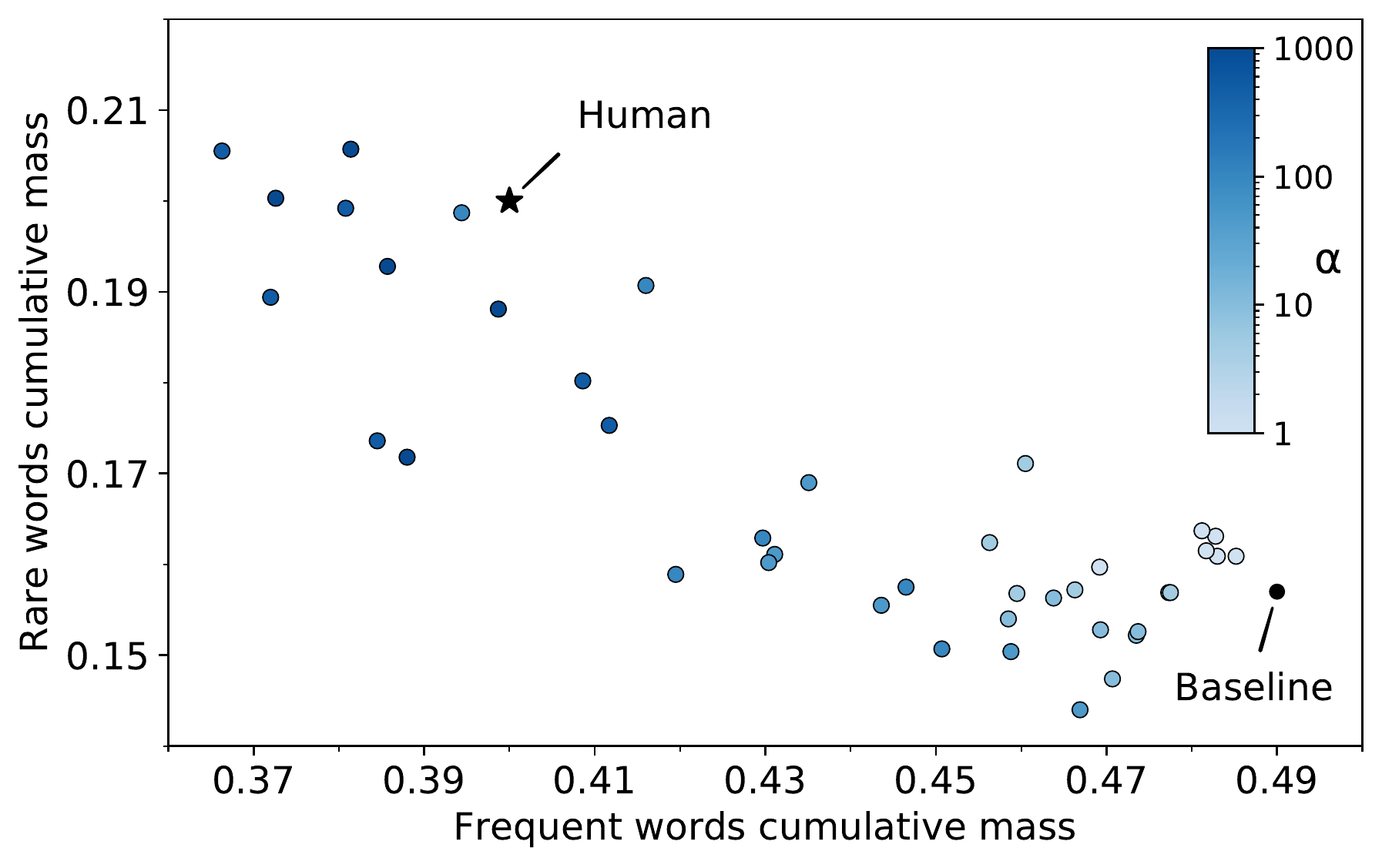}
    \caption{Vocabulary control with unlikelihood training: more probability mass is transferred from Frequent words to Rare words as we increase the $\alpha$ weighting parameter. The maximum likelihood baseline is far from the human distribution. }
    \label{fig:vocabalpha}
\end{figure}

Figure \ref{fig:vocabalpha} shows how the
vocabulary distribution obtained
after unlikelihood training
is affected by the choice of mixing hyperparameter 
$\alpha$ (Eq.~\ref{eq:mix}):
it can smoothly transition 
between the human training distribution and the  MLE trained distribution (`Baseline'), which is far from the human one.

Table \ref{tab:vocab} compares the MLE baseline with unlikelihood with increasing $\alpha$ values in terms of distribution and F1 score.
The vocabulary unlikelihood fine-tuning shifts probability mass from the over-represented frequent words towards under-represented medium and rare words, with the effect strengthening as $\alpha$ increases. At a small cost to perplexity and F1, the  unlikelihood tuning reduced the overuse of common tokens by 9 points, matching the human rate, while improving the production of rare tokens by 3 percentage points.

\if 0
Nucleus sampling is a popular method that can also produce generations closer to the human
vocabulary distribution. It does this by sampling from the model's probability distribution rather than using beam search, where the  sampler restricts  to the smallest set of tokens with total mass above a threshold $p\in[0,1]$. Small values of $p$ are similar to greedy sampling. Increasing $p$ yields distributions closer to human, but with large losses in F1 score, e.g. $p=0.5$ has a similar distribution to unlikelihood with 
$\alpha=10^2$ but the F1 scores are $0.160$ vs. $0.190$. This can be understood because
maximizing likelihood during decoding  yields better token accuracy than 
sampling \citep{welleck2019neural}, so the unlikelihood training approach to both use
likelihood decoding {\em and} match the human distribution can obtain the best of both worlds.
\fi

\paragraph{Human Evaluation}

\begin{figure}[t]
    \centering
    \includegraphics[width=\linewidth]{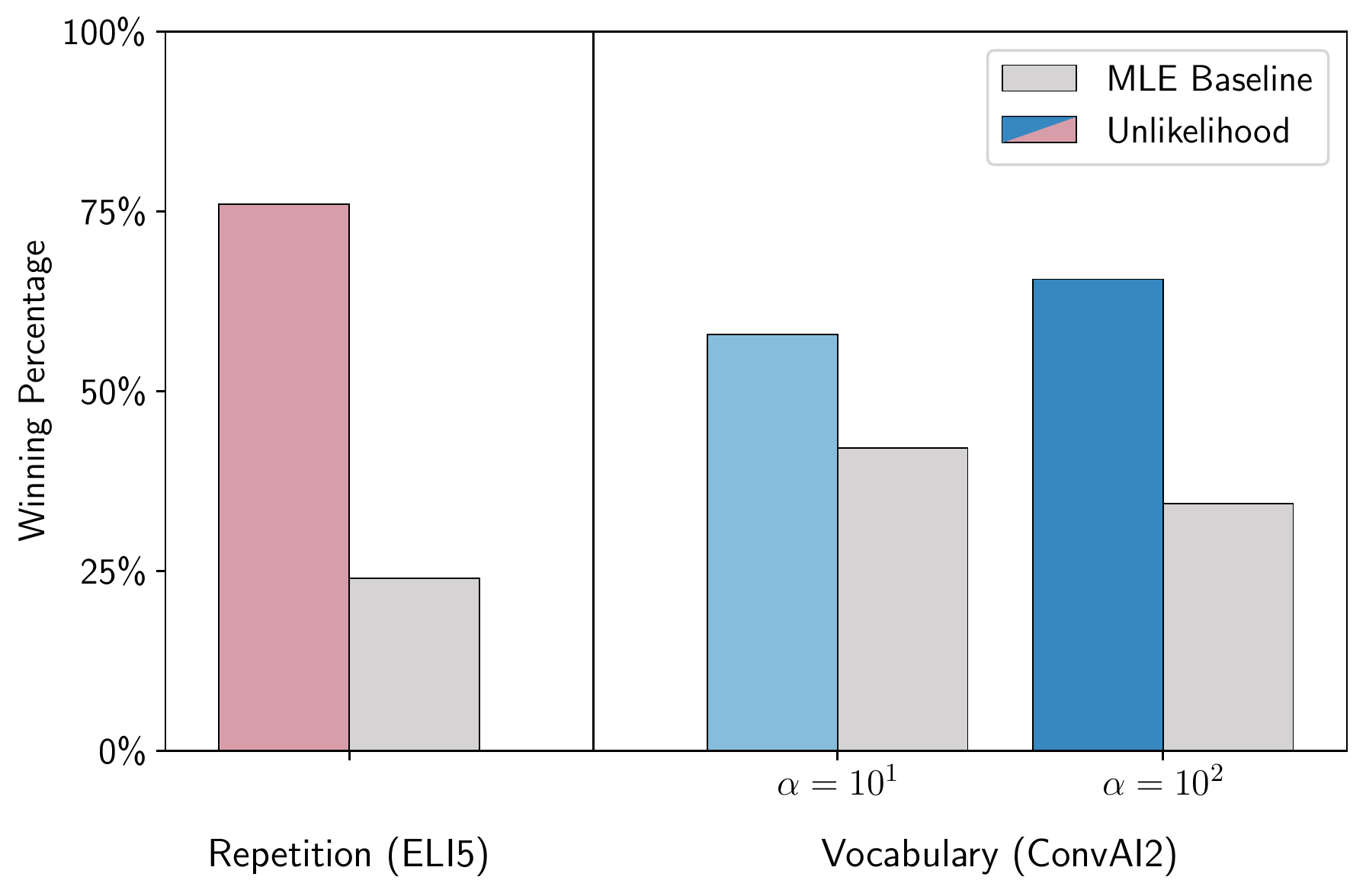}
    \caption{Human evaluation experiments for label unlikelihood on ELI5 (left), and vocabulary unlikelihood on ConvAI2 for two values of $\alpha$ (right). Unlikelihood significantly outperforms the MLE baselines.}
    \label{fig:human_eval}
\end{figure}

Finally, we perform a human evaluation using the ACUTE-EVAL framework \citep{li2019acute}, comparing the MLE baseline
to UL for various $\alpha$. First, 252 human-bot conversations (8 turns each) are collected, and then models are compared pairwise
by asking the question: {\em Who would you prefer to talk to for a long conversation?}
For these experiments we compare with both methods generating using beam with 
context blocking of trigrams.
 Results are given in Figure~\ref{fig:human_eval} (right), 
showing a statistically significant improvement over the baseline according to humans (two tailed binomial test, $p<0.01$). Further details are given in Appendix \ref{app:human_eval}.

\begin{table}[t!]
\setlength{\tabcolsep}{5pt}
    \centering
    \resizebox{\linewidth}{!}{
    \begin{tabular}{lrrrrrr}
\toprule
           &        &       &     \multicolumn{4}{c}{Token frequency classes}    \\
                        \cmidrule(lr){4-7}                                                                            
 Model       & PPL   & F1 & Freq & Med & Rare & Rarest \\
\midrule
 Human                   & -   & - & .400 & .300 & .200 & .100\\
 MLE Baseline            & 11.4 & .199 & .491     & .282   & .157 & .068  \\
\addlinespace[0.3em]
UL, $\alpha = 10^0$      &  11.4 & .200 &.483 &.289 & .163 & .063\\
UL, $\alpha = 10^1$      & 	11.9 & .201 &.459 &.328 & .154 & .058\\
UL, $\alpha = 10^2$      &  12.5 & .190 &.430 &.335 & .163 & .071\\
UL, $\alpha = 10^3$      &  14.4 & .174 &.399 &.339 & .188 & .073\\
 \addlinespace[0.3em]
\bottomrule
\end{tabular}
    }
    \caption{Unlikelihood loss applied to vocabulary distributions. Stronger $\alpha$ terms greatly shift probability mass from the most Frequent words to Medium and Rare words, at a small cost to PPL and F1.  
    Frequent, medium, rare and rarest token classes are defined as the sets of tokens whose cumulative masses account for the top 40\%, the next 30\%, the next 20\% and final 10\% of tokens empirically generated by humans, respectively.}
    \label{tab:vocab}
\end{table}

\subsection{Contradictions}

\begin{figure}[t]
\centering
\begin{minipage}{.5\textwidth}
  \includegraphics[width=0.95\linewidth]{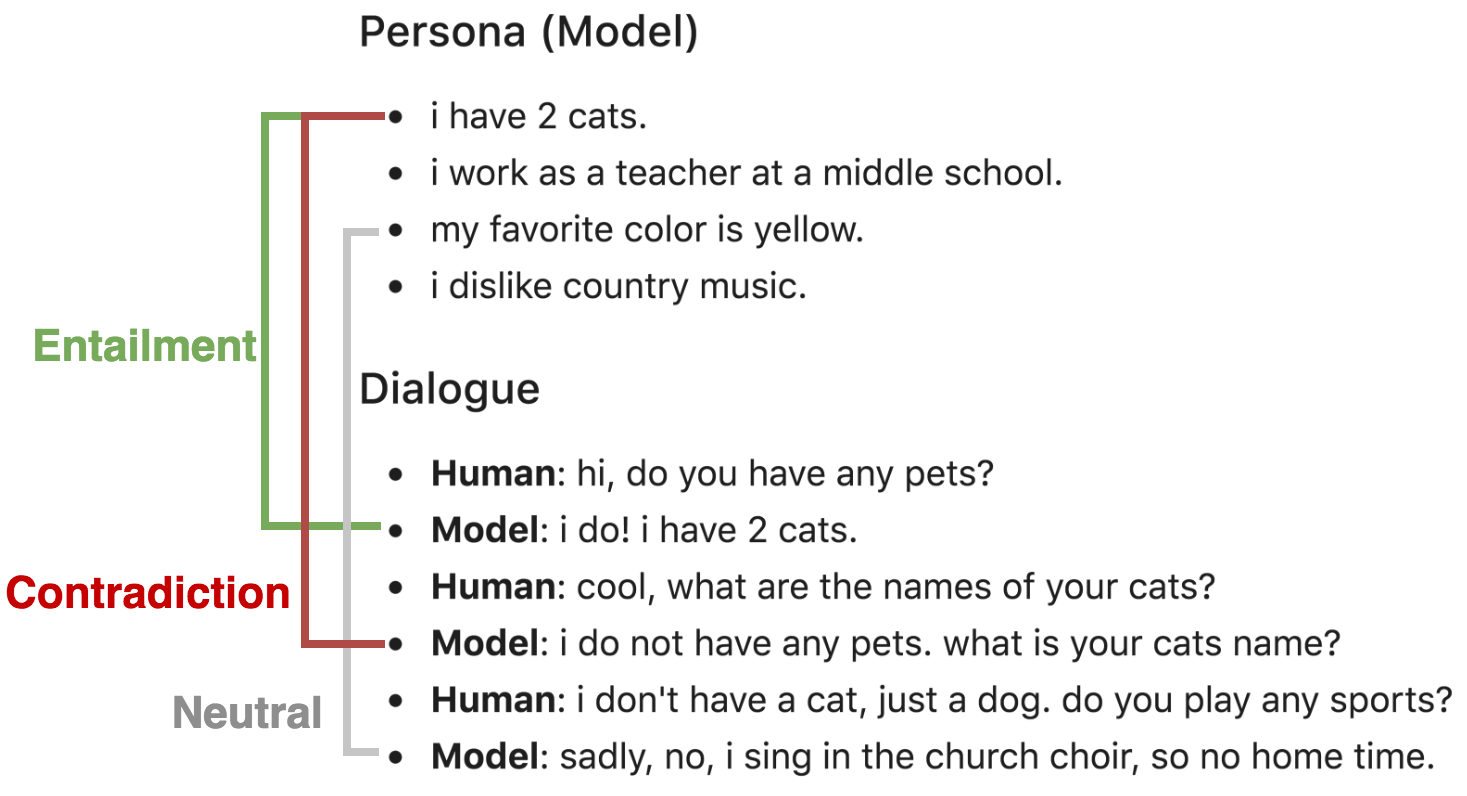}
  \captionof{figure}{Dialogue NLI from \citep{welleck2018dialogue}.}
  \label{fig:dnli}
\end{minipage}
\end{figure}

\begin{table}[t]
   \centering
    \begin{tabular}{lrrr}
\toprule
                 & Train & Test & Valid\\
\midrule
Entailment        & 95k   & 4613  & 4959    \\
Triple-Entailment & 105k  & 5285  & 5481    \\
Neutral           & 110k  & 5500 & 5700     \\
Negatives         & 110k  & 5500 & 5700    \\
\bottomrule
    \end{tabular}
    \caption{Dialogue NLI two utterance generation task dataset statistics.
 }    \label{tab:nli_2sentence_stats}
\end{table}

\begin{table*}[t]
    \centering
    \begin{small}
    \begin{tabular}{lrrrrrrrr}
\toprule
                 & \multicolumn{3}{c}{Selection Accuracy} &   \multicolumn{5}{c}{Perplexity} \\
\cmidrule(lr){2-4} \cmidrule(lr){5-9}
Data + Model       &  Entail  &  Tr.-E &  Neutral  &   Entail & Tr.-E & Neutral & Contradict & ConvAI2\\
\midrule
MLE Baseline      &  72\% & 41\% & 18\% &  8.54 &  17.5 & 36.7  & 12.5 & 11.4 \\
UL (Dialogue NLI) &  96\% & 85\% & 78\% &  9.1	 &  26.6 & 39.4  & 248.9 & 11.9\\                                                                                         
\bottomrule
    \end{tabular}
    \end{small}
    \caption{Test evaluation on the Dialogue NLI two utterance generation task, comparing standard likelihood (MLE)
models trained on pushshift.io Reddit and ConvAI2 with unlikelihood loss NLI training. Results are broken down according to whether the premise and positive candidate are entailing, triple-entailing, or neutral (Entail, Tr.-E, Neutral). Selection Accuracy measures how often the model assigns lower perplexity to the positive candidate than to the negative candidate in the pair. Top two rows: for standard maximum likelihood models, the perplexity of contradicting utterances is {\em lower}  compared to neutral or triple-entailing utterances (albeit higher compared to entailing utterances), showing partial failure at the coherence task. Bottom row: NLI Unlikelihood training yields large improvements on all coherence metrics, while minimally increasing overall perplexity.
 }
    \label{tab:nli_2sentence_hits}
\end{table*}

\begin{table*}
    \centering
    \begin{small}
    \begin{tabular}{lrrrrrr}
\toprule
                 & \multicolumn{2}{c}{Selection Accuracy (vs. Neg)} &   \multicolumn{4}{c}{Perplexity} \\
\cmidrule(lr){2-3} \cmidrule(lr){4-7}
Data + Model       &    Triple-Entail  &  Neutral  &   Triple-Entail & Neutral & Contradict & ConvAI2\\
\midrule
 MLE Baseline         &  66.5\%       & 36.8\% &  23.3  & 45.1  & 35.9  & 11.4 \\
UL (Dialogue NLI) &  89.0\% & 69.8\% &  21.5  & 40.3  & 63.5  & 11.8\\
\bottomrule
    \end{tabular}
    \end{small}
    \caption{Test evaluation on the Full Dialogue NLI generation task. 
    NLI unlikelihood training improves coherence metrics compared to likelihood (MLE) training. For UL, the triple-entailing or neutral candidates are assigned relatively lower perplexity compared to contradicting candidates, with higher selection accuracy for coherent labels. 
 }
    \label{tab:nli_full_hits}
\end{table*}
\begin{figure*}[h!]
    \centering
    \begin{small}
    \begin{tabular}{llrr}
\toprule
                                             &            & $\mathcal{L}_{\text{MLE}}$ & $\mathcal{L}_{\text{UL}}$ \\
Premise                                      & Hypothesis & PPL & PPL \\
\midrule
Yes, I love watching baseball   and basketball. I do not 
  & (C) I love running.    & 25.5 & 226.9 \\ 
like running though.&  (E) I despise running. & 29.9 & 9.4 \\
\midrule
Yes, I love watching baseball and basketball.  I do like 
 & (E) I love running.    & 26.2 & 3.1 \\ 
running though. & (C) I despise running. & 42.8 & 247.1 \\
\midrule
We did too but working in real estate for 12 years 
.  &  (E) I have been working as a real estate\\ 
sucked up a lot of time& agent for the past 12 years.      & 3.9 & 3.8  \\
& (C) We did too but working in real estate\\
& for fifteen years sucked up a lot of time. & 3.1 & 17.6\ \\
\bottomrule    
\end{tabular}
    \end{small}
    \caption{Example perplexities of a baseline maximum likelihood model ($\mathcal{L}_{\text{MLE}}$) and our unlikelihood trained model ($\mathcal{L}_{\text{UL}}$ ) when 
    generating the provided hypotheses, given the premise. The maximum likelihood trained
     model assigns high probability (low perplexity) to contradictory generations, while unlikelihood does not.}
    \label{tab:nli_ex}
\end{figure*}

We use the dialogue natural language inference (NLI) task of \citet{welleck2018dialogue} to obtain labeled non-contradicting and contradicting dialogue sentence pairs to use in unlikelihood training (\S\ref{ssec:contra-ul}).
Dialogue NLI contains utterances labeled as 
entailing (E), neutral (N) or contradiction (C), given a premise that is either a persona sentence (an initial context sentence describing a dialogue agent's personality) or another dialogue utterance
from the Persona-Chat dialogue task \citep{zhang2018personalizing}.
We show examples from Dialogue NLI in Figure \ref{fig:dnli}.
The original data
consists of sentence pairs $(s_1,s_2)$
along with a label (E, N, or C),
and was constructed by developing a schema and employing crowdworkers to label
utterances with relation triples.  The labels are
then inferred from the triple representation.

We first transform the original classification dataset into a form useful
for unlikelihood training of a generative dialogue model.
We consider two setups: (i) a two utterance generation task; and 
(ii) a full dialogue generation task.

\if 0
(i) a two utterance generation task, whereby given a single utterance as context, the task is to generate a non-contradicting second utterance, 
and (ii) a full dialogue generation task, where the context is a full persona and dialogue history instead.
\fi

\paragraph{Two Utterance Generation Task}
We adapt the initial dialogue NLI dataset by using entailing and neutral training sentence pairs as plausible positive utterances, and contradicting pairs as negatives. That is, if a pair $(s_1,s_2)$ from Dialogue NLI has label E or N,  the example $(\mathbf{x},\mathbf{y})=(s_1,s_2)$ is added to $\mathcal{D}^+$, otherwise (label C) it is added to $\mathcal{D}^-$.

We consider two types of entailment: entailing sentence pairs that appear together in a dialogue in the original Persona-Chat dataset and are therefore natural (`entailment'), and those that only 
entail via their triple relations (`triple-entailment'). The latter are more challenging, noisier targets.
Evaluation is performed by measuring the test set
perplexity over the four target 
label types, where contradictions should have relatively higher perplexity. 
We additionally evaluate a selection accuracy task, where for each test example there
are two candidate responses:  a positive and a negative (contradicting) statement. 
The candidate response with the lowest perplexity is considered to be the model's selection, and we measure the selection success rate.  Evaluation is broken down by positive type (entailment, triple-entailment, neutral).
Dataset statistics are given in Table \ref{tab:nli_2sentence_stats}.

\paragraph{Full Dialogue Task}
To evaluate in a more realistic setup that involves full dialogue rather than a single utterance, we take full Persona-Chat dialogues \citep{zhang2018personalizing} similar to Figure \ref{fig:dnli},  and map back the dialogue NLI data to provide positive and negative continuations of the dialogue. We consider continuations as either triple entailing utterances,  neutral utterances or contradictions --  where the relation triple is used to match the existing persona or dialogue turns by the same speaker to induce the label. That is, an example $(\mathbf{x},\mathbf{y})$ consists of a dialogue history $\mathbf{x}=\{p_1,\ldots,p_k,u_1,\ldots,u_t\}$ and utterance $\mathbf{y}=s_2$, where $(s_1,s_2)$ is a sentence pair from Dialogue NLI, and at least one sentence in $\mathbf{x}$ has the same relation triple as $s_1$. When the pair $(s_1,s_2)$ is labeled as E or N in Dialogue NLI, the example $(\mathbf{x},\mathbf{y})$ is added to $\mathcal{D}^+$, and otherwise it is added to $\mathcal{D}^-$.

\paragraph{Results}
\if 0
We first report metrics for our baseline likelihood (MLE) pre-trained
pushshift.io Reddit model,
and the same model fine-tuned with MLE on the Persona-Chat ConvAI2 dataset.  The reason
to fine-tune on a well-known existing dataset is to measure
if the dialogue model we are using is
matching the state-of-the-art, independent
of measures of coherence.
We obtain a perplexity of 11.4 with
the fine-tuned model, which is line with current best systems on this task
\citep{lewis2019bart}.
\fi 

Our MLE baseline obtains a perplexity of 11.4, 
in line with current best systems on this task \citep{lewis2019bart}.
Unfortunately, despite being good on such standard metrics, our baseline
models fail at our coherence task. 
As seen in Table \ref{tab:nli_2sentence_hits} for the two utterance task, the perplexity of contradicting utterances (12.5) is on average {\em lower} than for neutral (36.7) or triple-entailing utterances (17.5), although it is higher than entailing utterances. 
We believe this is due to contradicting utterances having high word overlap with the premise utterance, 
coupled with an inability to judge incoherence.
Viewed as a selection task between utterances, picking the utterance with the lowest perplexity, this means the selection rates of non-contradicting utterances are very low, e.g. 
picking neutral utterances over contradicting utterances only 18\% of the time.
Even fully entailing utterances are only picked 73\% of the time.
Similar results are found on the full dialogue task as well, see Table \ref{tab:nli_full_hits}.

Unlikelihood training brings large improvements in coherence metrics, whilst minimally impacting overall dialogue perplexity.
After applying unlikelihood, perplexity for contradicting utterances has a clear signature, with very large average values compared to entailing or neutral utterances, e.g. 248.9 vs. 9.1 for contradict vs. entail on the two utterance task. This converts to corresponding large increases in selection accuracy across all types on both tasks, e.g., an increase from 18\% to 78\% on neutral statements on the two utterance task, and from 37.4\% to 69.8\% on the full dialogue task.

Some example model predictions are given in Figure \ref{tab:nli_ex}, comparing the MLE baseline and
  unlikelihood model  perplexities of generating the given hypotheses.
The likelihood model cannot differentiate between contradicting and entailing statements easily, while there are large perplexity differences for the unlikelihood model in these cases.

\section{Conclusion}
Generating
 consistent and coherent human-like dialogue is a core goal of natural language research. We studied several aspects that contribute to that goal, defined 
metrics to measure them, and proposed algorithms that improve them, 
mitigating some of 
the failings of  maximum likelihood training, the current dominant approach.
Our method defines
objective functions under the umbrella of 
unlikelihood: during training, we wish to make inconsistent dialogue
unlikely by lowering the probability of such events occurring.
This makes generative models repeat themselves less,  
copy the context less, and use more rare words from the vocabulary -- closer to
matching human statistics. Further, utilizing supervised datasets with labeled 
coherent and incoherent utterances and applying unlikelihood yields 
measurably improved levels of coherence 
with respect to the aspect measured, in this
case contradiction.
Future work could apply this same technique with other supervised data, 
e.g. correcting causal or commonsense reasoning errors
\citep{zellers2019hellaswag,qin2019counterfactual}.

\noindent
\bibliography{bib}
\bibliographystyle{acl_natbib}

\newpage
\appendix

\clearpage

\section{Repetition Control with Beam Search} \label{app:beam}

The experiments on repetition and copying in the main paper
were carried out with greedy decoding for simplicity.
In this section we show that similar results hold with beam decoding as well.
Using a beam size of 5, we take the same 4 models from Table \ref{tab:rep_wiz}
and compute metrics with beam instead. 
The results are given in Table \ref{tab:rep_wiz_beam} which show similar trends to before,
except the baseline model using beam tends to suffer more from repetition, which is a 
known result \cite{holtzman2019curious}.
Note that we simply evaluated the same unlikelihood models as before, but we expect
that better results could be obtained by performing sequence level unlikelihood training
with beam search in the training loop, as well as choosing hyperparameters specifically
with this kind of decoding being used to measure validation performance.

\begin{table}
\setlength{\tabcolsep}{4pt}
    \centering
    \resizebox{0.95\linewidth}{!}{
    \begin{tabular}{lrrrr}
\toprule
     &     &   &  \multicolumn{2}{c}{Repetition} \\
     \cmidrule(lr){4-5}
 Model              & PPL  & F1  & Context & Label \\
\midrule
Human                 & -    &  -   & .160 & .0006  \\
MLE Baseline          & 8.3  & .373 & .582 & .002 \\
\addlinespace[0.3em]
UL (Context only)      & 8.8  & .345 &  .270  & .001  \\
UL (Label only)        & 8.3  & .371 &  .645  & .000  \\
UL (Context + Label)  & 8.5  & .358 &  .445  & .003   \\
\bottomrule    
\end{tabular}
    }
    \caption{Evaluation on the Wizard of Wikipedia task test set, comparing standard likelihood (MLE) with repetition unlikelihood loss training, where both methods use beam search (beam size of 5).
}
    \label{tab:rep_wiz_beam}
\end{table}

\section{Nucleus Sampling for Vocabulary control}

\begin{table}[t!]
\setlength{\tabcolsep}{5pt}
    \centering
    \resizebox{\linewidth}{!}{
    \begin{tabular}{lrrrrrr}
\toprule
           &        &       &     \multicolumn{4}{c}{Token frequency classes}    \\
                        \cmidrule(lr){4-7}                                                                            
 Model       & PPL   & F1 & Freq & Med & Rare & Rarest \\
\midrule
 Human                   & -   & - & .400 & .300 & .200 & .100\\
 MLE Baseline            & 11.4 & .199 & .491     & .282   & .157 & .068  \\
\addlinespace[0.3em]
Nucleus $p=0.3$ & 11.4 & .180 & .452 & .315 & .168 & .064 \\
Nucleus $p=0.4$ & 11.4 & .171 & .440 & .320 & .172 & .068\\
Nucleus $p=0.5$ & 11.4 & .160 & .425 & .322 & .180 & .072\\
Nucleus $p=0.6$ & 11.4 & .151 & .411 & .318 & .192 & .078\\  
Nucleus $p=1.0$ & 11.4 & .141 & .394 & .302 & .201 & .101\\
\addlinespace[0.3em]
UL, $\alpha = 10^0$      &  11.4 & .200 &.483 &.289 & .163 & .063\\
UL, $\alpha = 10^1$      & 	11.9 & .201 &.459 &.328 & .154 & .058\\
UL, $\alpha = 10^2$      &  12.5 & .190 &.430 &.335 & .163 & .071\\
UL, $\alpha = 10^3$      &  14.4 & .174 &.399 &.339 & .188 & .073\\
 \addlinespace[0.3em]
\bottomrule
\end{tabular}
    }
    \caption{Unlikelihood loss applied to vocabulary distributions. Stronger $\alpha$ terms greatly shift probability mass from the most Frequent words to Medium and Rare words, at a small cost to PPL and F1.  
    Frequent, medium, rare and rarest token classes are defined as the sets of tokens whose cumulative masses account for the top 40\%, the next 30\%, the next 20\% and final 10\% of tokens empirically generated by humans, respectively.   Nucleus sampling can also produce a distribution close to human with parameter $p$ close to 1, but with larger losses in F1. }
    \label{tab:vocab2}
\end{table}

Table \ref{tab:vocab2} compares the MLE baseline, unlikelihood with increasing $\alpha$ values, and Nucleus sampling \cite{holtzman2019curious} with hyperparameter $p$ in terms of distribution and F1 score.
The vocabulary unlikelihood fine-tuning shifts probability mass from the over-represented frequent words towards under-represented medium and rare words, with the effect strengthening as $\alpha$ increases. At a small cost to perplexity and F1, the  unlikelihood tuning reduced the overuse of common tokens by 9 points, matching the human rate, while improving the production of rare tokens by 3 percentage points.

Nucleus sampling is a popular method that can also produce generations closer to the human
vocabulary distribution. It does this by sampling from the model's probability distribution rather than using beam search, where the  sampler restricts  to the smallest set of tokens with total mass above a threshold $p\in[0,1]$. Small values of $p$ are similar to greedy sampling. Increasing $p$ yields distributions closer to human, but with large losses in F1 score, e.g. $p=0.5$ has a similar distribution to unlikelihood with 
$\alpha=10^2$ but the F1 scores are $0.160$ vs. $0.190$. This can be understood because
maximizing likelihood during decoding  yields better token accuracy than 
sampling \citep{welleck2019neural}, so the unlikelihood training approach to both use
likelihood decoding {\em and} match the human distribution can obtain the best of both worlds.

\section{Human Evaluation}
\label{app:human_eval}

\begin{figure*}[ht!]
    \centering
    \includegraphics[width=\linewidth]{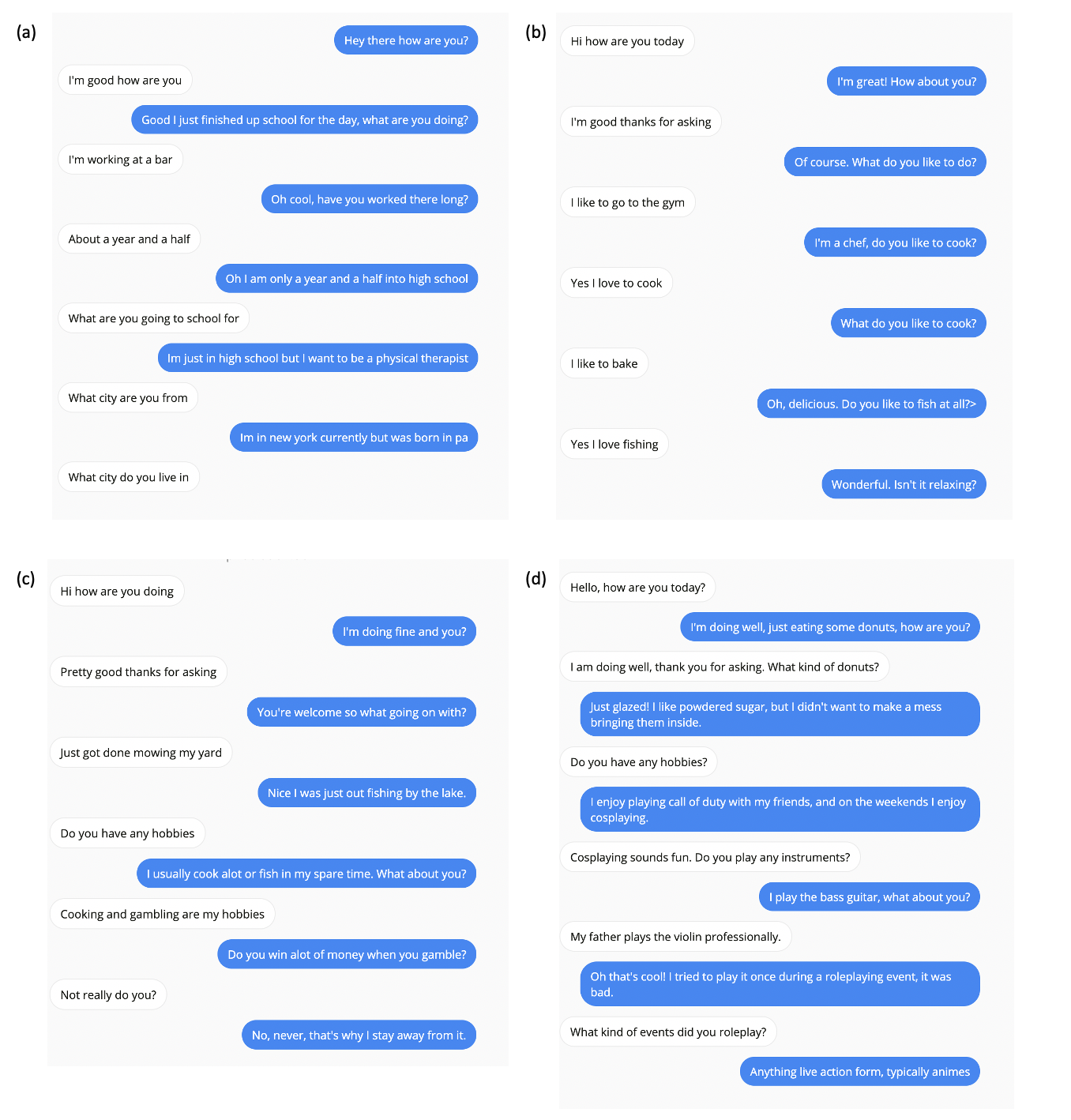}
    \caption{Examples of model-human conversations collected during human evaluation of the vocab unlikelihood models. Human utterances are in blue bubbles, model utterances are in white. Conversations (a) and (b) are from the baseline. Conversations (c) and (d) are from the $\alpha=10^2$ model and more frequently employ rarer words.}
    \label{fig:vocab_examples}
\end{figure*}

\paragraph{Description of ConvAI2 vocabulary setup}
We follow \citep{li2019acute} and perform a pairwise comparison with full-length model conversations. We first collected 252 model-human conversations with each of the models (MLE baseline, and weights for $\alpha$ of Unlikelihood, examples in \ref{fig:vocab_examples}). We then set up a pairwise-comparison using the software of \citep{li2019acute}, using the same question (``{\em Who would you prefer to talk to for a long conversation?}'') and use the exact same quality control question (a baseline greedy model without repetition control, versus a human). We collected approximately 200 preferences per model comparison and filtered annotators who failed quality control.

\begin{figure*}[ht!]
    \centering
    \includegraphics[width=\linewidth]{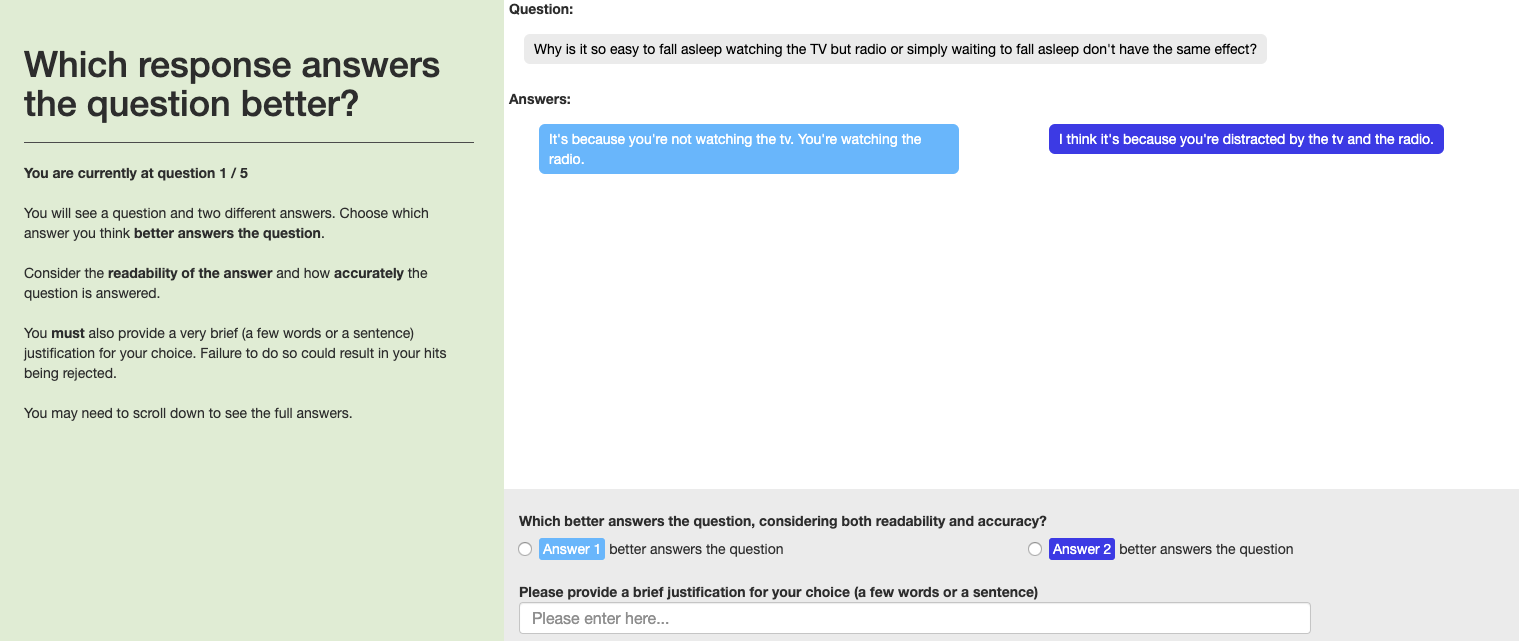}
    \caption{Screenshot of the Human Evaluator UI.}
    \label{fig:eli5_ss}
\end{figure*}

\paragraph{Description of ELI5 repetition setup}
We follow \citep{fan2019eli5} and perform a pairwise evaluation where human annotators were asked ``{\em which response answers the question better?}'' A screenshot of the UI is shown in Figure~\ref{fig:eli5_ss}. Human evaluators were asked to rate a total of 5 questions, two of which were quality control annotations. The quality control examples contained the real human responses, along with model predictions: one question contained a baseline model, and one contained an unlikelihood model. Annotators which did not pick humans in quality controls were removed from the final setups. We collected 200 annotations comparing the baseline and the unlikelihood model.

\paragraph{Results}
\begin{figure}[t]
    \centering
    \includegraphics[width=\linewidth]{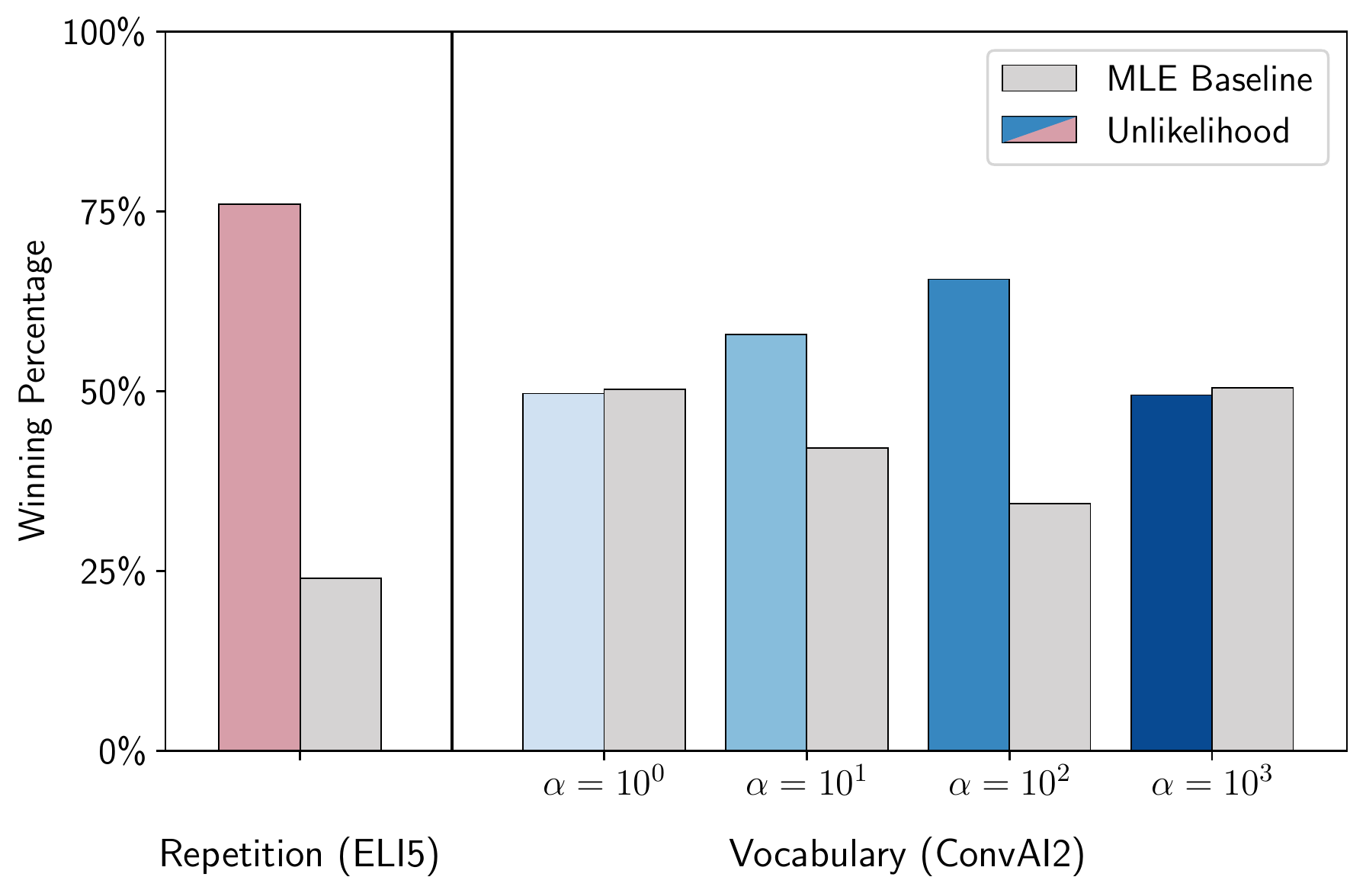}
    \caption{Complete Human Evaluation results. Human evaluators do not significantly prefer the $\alpha=10^0$ and $\alpha=10^3$ models over the baseline model.}
    \label{fig:extended_human}
\end{figure}

Evaluation results from all evaluated matchups are shown in Figure~\ref{fig:extended_human}. We find our repetition-controlled ELI5 model significantly outperforms the MLE baseline. We find that two of the vocabulary repetition significantly outperform the MLE baseline. We compute significance with a two-tailed binomial test ($p < .01$).

\end{document}